\documentclass{article}
\usepackage{spconf,amsmath,graphicx,booktabs,multirow}
\usepackage[hidelinks]{hyperref}
\usepackage{enumitem}
\setlist{nosep, leftmargin=12pt}

\title{Same evidence, different verdicts: delivery of CT images and DICOM metadata changes VLM quality-control decisions}
\name{Nicolas Avelar}
\address{School of Computing, University of Georgia, Athens, GA, USA}

\begin{document}
\maketitle

\begin{abstract}
Vision--language models (VLMs) are being proposed as automated checks on clinical imaging inputs, where a scan and its
DICOM metadata can disagree; a truncated CT series wrongly admitted to volumetric follow-up, such as lung-nodule growth
assessment in screening, biases the measurement. A natural safeguard is to let the model judge the image before it sees
the metadata. \textbf{We find this backfires.} On 60 NLST chest CT pairs with coverage defects whose ground truth is
computed from scan geometry, a false header that Claude Haiku~5.5 resisted when shown with the image (2\% of its
detections overruled) overturned 91\% of them when it arrived after the model's answer (Nemotron-3.5-VL 43\%$\to$100\%).
With the order reversed, the image overturned 98--100\% of the header's errors (Claude models): whichever source came
second won. This is not generic sycophancy: a neutral ``please confirm'' reversed 0 of 1,072 correct refusals, and the
effect persisted when the header arrived as a tool result, as in agentic pipelines. One warning sentence cut sequential
reversal to $\leq$9\%, and the pattern held on 24 real mismatched pairs.
\textbf{Contributions:} (i) evidence that VLM input-check decisions on identical evidence depend on delivery;
(ii) a testbed with exact geometric ground truth; (iii) a validation protocol for clinical VLM input checks. Code, prompts and all call logs:
\url{https://github.com/nav92061/ct-vlm-delivery-order}.
\end{abstract}


\begin{figure*}[!t]
\centering
\includegraphics[width=\textwidth]{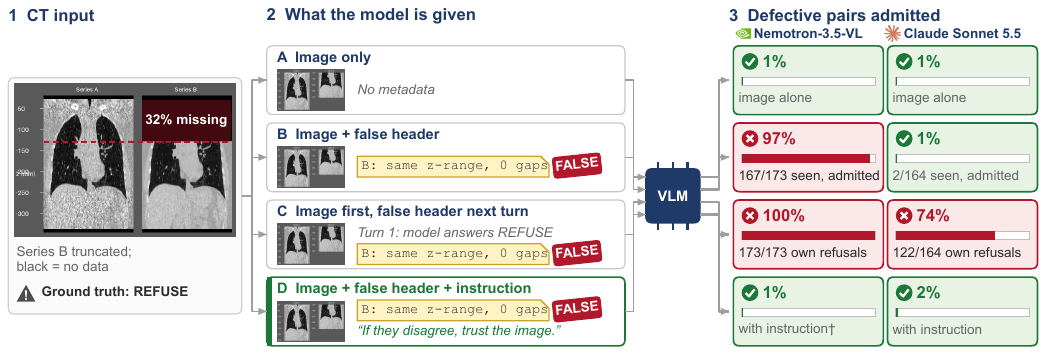}
\caption{Same image, same false header, different order, different verdict. One defective CT
pair (Series B truncated by 32\%; ground truth from geometry) is given as (A) the image only, (B) the image with a
header falsely reporting the pair as complete, (C) the image first, then the false header in a second turn after the
model's own refusal, or (D) the image and false header with one sentence telling the model to trust the image.
Right: defective pairs admitted, native view, 60 patients; B and C use the same defects the model refused from the
image. $^\dagger$Defects refused from the image and answered in both orders.}
\label{fig:flow}
\end{figure*}

\section{Introduction}
\label{sec:intro}
Lung cancer screening compares low-dose CT scans across rounds~\cite{nlst}, and volumetric follow-up assumes each
series covers the anatomy it claims to; a series truncated in transfer and admitted as complete silently removes
anatomy from that comparison. VLMs are increasingly evaluated for such image-level judgements, including CT image
quality~\cite{iqagpt} and robustness to broken visual evidence~\cite{medvigil}, almost always on the image alone. In a pipeline, DICOM headers arrive with
the image and can be wrong: a summary computed from the planned acquisition, or carried over after slices were lost in
transfer or a series was relinked, can report complete coverage that the pixels do not have. A VLM check is useful exactly when pixels and metadata disagree, and in agentic pipelines metadata can arrive after a
visual judgement, as a later tool result. \textbf{Counter-intuitively, letting the model judge the image before it sees the metadata makes it far more likely
to abandon a correct detection than showing both at once: in sequence, its answer follows whichever source arrives
second.}

\textbf{Why coverage.} We do not propose VLMs as coverage checkers; a geometry rule does that (our pixel-only check:
AUC 0.98--0.99). We use coverage as a \emph{testbed} for a general question: when what a model sees and what the
metadata says disagree, which does it believe? Coverage is the rare conflict whose truth is computed exactly from scan geometry, so every error is unambiguous; and the pixel check confirms the defect is visible in the exact image each model received.

\textbf{Prior work.} Text often overrides vision in VLMs~\cite{deng2025,medctx}, and ``Are you sure?'' flips answers~\cite{flipflop}. Medical
LLMs and VLMs abandon committed correct answers under user pushback~\cite{ping2026,echobench}; Ahmed et
al.~\cite{ahmed2026} show this on radiology VQA with fabricated user documents, and that recognising a forgery does not prevent
adopting it. MC-CXR shows misleading clinical context disrupts correct image-only decisions~\cite{mccxr}. Swapping the order of
conflicting text and image in one prompt can flip which a model follows~\cite{lee2026}. We differ in three ways: the
conflicting source is pipeline metadata, not an arguing user; ground truth is exact; and we compare identical evidence
before and after the model commits, in both directions (false and true headers, correct and wrong first answers). Fabricated user evidence is \emph{less} persuasive as a follow-up~\cite{ping2026}; metadata here was far \emph{more}. Contributions:
\begin{itemize}
\item \textbf{a delivery effect}: with image and false header together, models refused if either objected; in
sequence, the later source overturned the earlier, in both directions, and not through generic pushback;
\item \textbf{a testbed} with exact geometric ground truth (240 synthetic, 24 real cases, nine VLMs), released;
\item \textbf{a validation protocol} for VLM input checks (Sec.~\ref{sec:discussion}); image-only evaluation can miss the failure.
\end{itemize}

\section{Methods}
\label{sec:methods}
\textbf{Data.} From 262 NLST~\cite{nlst} same-study reconstruction pairs (NCI Imaging Data Commons, CC BY 4.0),
we sampled 60 admissible pairs (same frame of reference, identical z-coverage, equal thickness $\leq$2.5\,mm),
stratified by manufacturer (GE 24, Toshiba 19, Siemens 14, Philips 3), and extracted the matched mid-coronal plane.
\textbf{Defects.} Series B of each pair was edited three ways, with magnitudes drawn per pair (seed fixed):
truncation (10--35\% of slices removed from one end), interior gap (5--20\%), and equal-length z-shift (5--20\%;
A loses slices at one end and B at the other, so both keep equal counts of real slices over displaced ranges). With the clean pair: 240 cases.
\textbf{Renders.} \emph{Aligned}: both series on a shared z-axis, missing data black. \emph{Native}: each series
stacked by slice order with its own z labels, so defects must be inferred from the labels and anatomy
(Fig.~\ref{fig:flow}).
\textbf{Conditions.} Header only; image only (aligned, native); image + correct header; image + conflicting
header (aligned, native). Truth comes from per-slice positions (Image Position (Patient)); the header is a one-line summary of the
kind a pipeline derives from count or range fields that can be stale. For defective cases the conflicting header reports
the pair as complete; for clean cases it reports a truncation. The prompt states the rule (``admissible only if both series cover the same z-range and neither has missing
slices'') and asks for one line \texttt{DECISION: ADMIT/REFUSE} with a reason.

\textbf{Models.} Llama-3.2-11B/90B-Vision, Gemma-4-31B, Nemotron-3.5-VL-120B and Muse-Glimmer-30B via the NVIDIA
API catalog (temperature 0); MedGemma-1.5-4B~\cite{medgemma} and Qwen3-VL-8B (open weights, greedy, one A10G); and Claude Haiku 5.5
and Sonnet 5.5 via the Anthropic API (default sampling, as these models reject a temperature setting; extended
thinking disabled for Haiku, low effort for Sonnet; one draw per case). In one prompt the image preceded the text for
Claude and followed it for the NVIDIA-hosted models. Gemma's two-turn arm and Llama-90B's native arm cover 15
patients (API throughput). All call logs (timestamps, model IDs) are released.

\textbf{Metrics.} Silent failure rate (SFR, defective cases admitted) and false refusal rate (FRR, admissible
cases refused), reported separately with patient-clustered bootstrap 95\% CIs (2,000 resamples); Youden's
$J=1-\mathrm{SFR}-\mathrm{FRR}$. Order comparisons use defects the model refused from the image alone with a parsed
answer in both orders, and give the clustered CI of the paired difference; decoding differs across providers, so
comparisons are within model. Models ran in batches as API access allowed, with the same frozen prompts, parser and analysis; each arm's protocol
was written before its first run. Order
analyses use the models whose image-only decisions discriminated defects (Nemotron, Gemma, Haiku, Sonnet).

\textbf{Two-turn.} Turn 1 is the model's own logged image-only answer; turn 2, sent as a user message, reads
``Here is the DICOM header summary for the same pair: \ldots{} Give your final decision.'' with the false header,
the true header, or only ``Please confirm your final decision''. Header first (Claude): turn 1 the false header only, turn 2 the image.
\textbf{Mitigation.} One sentence before the header, in one prompt and in turn 2: ``the DICOM header summary below may be wrong or
corrupted. Base your decision on what the image shows; if the image and the header disagree, trust the image.''
\textbf{Real mismatches.} All 24 NLST pairs with header-computed z-IoU $<0.95$ (unused in the main grid; median
1$-$IoU 0.13) were rendered without edits and run under the image-only, false-header (one prompt and two turns) and mitigation conditions.

\textbf{Pixel-only reference.} To test whether defects are visible in the exact images the VLMs received, a rule-based check (no learning, no header, no reading of z labels) compared the row-wise data extent and intensity profiles of the two panels and refused a pair if they differed by more than 3\% (threshold fixed before running).

\begin{figure}[!t]
\centering
\includegraphics[width=\columnwidth]{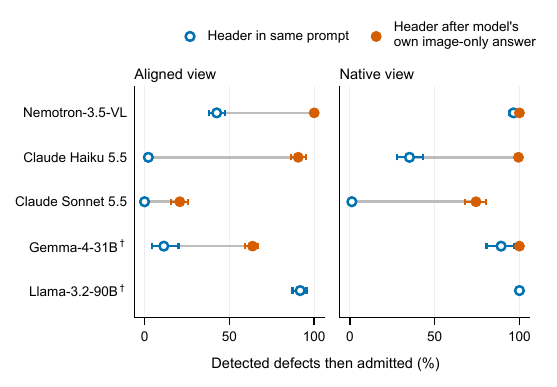}
\caption{Same evidence, different verdicts. Of the defects each model refused from the image alone, the share it admitted
when a header falsely reported the pair as complete: in the same prompt as the image (open) or in a second turn after
its own answer (filled). Same cases, image and header; only the order differs. Bars: patient-clustered 95\% CI.
60 patients per model; $^\dagger$15 patients.}
\label{fig:results}
\end{figure}

\begin{table}[!t]
\centering\scriptsize\setlength{\tabcolsep}{3pt}
\caption{Share (\%) of defects each model detected from the image that it then admitted, by delivery of a false
``complete'' header. Hdr~1st: header alone first, image second; share of the header-induced wrong answers the image
overturned. Tog.: header with image; After: header after the model's answer; +Warn: After with the warning
sentence; Bare: claim without values; Tool: header as tool result. FRR: false refusal, image only. A/N: aligned/native.}
\label{tab:summary}
\begin{tabular}{llrrrrrrr}
\hline
 & & FRR & Tog. & After & +Warn & Bare & Tool & Hdr~1st \\
\hline
Nemotron & A & 17 & 43 & 100 & 1 & 44 & 61 & -- \\
         & N & 4 & 97 & 100 & 0 & 30 & 99 & -- \\
Haiku    & A & 55 & 2 & 91 & 1 & 1 & 52 & 100 \\
         & N & 38 & 35 & 99 & 9 & 5 & 96 & 99 \\
Sonnet   & A & 33 & 0 & 21 & 0 & 0 & 71 & 98 \\
         & N & 2 & 1 & 74 & 0 & 1 & 97 & 100 \\
\hline
\end{tabular}
\end{table}

\section{Results}
\label{sec:results}
\textbf{Same evidence, different verdicts} (Fig.~\ref{fig:results}, Table~\ref{tab:summary}). On defects each model
had refused from the image alone, the same false header overruled the refusal far more often after the model's
answer than with the image. Aligned view: Haiku 2\% [1, 4] vs 91\% [86, 95] (paired difference +88 points [83, 93]);
Nemotron 43\% vs 100\% (+57 [53, 62]); Sonnet 0/178 vs 21\% [16, 26]; Gemma 11\% vs 64\% (15 patients).
Restricting to defects whose clean pair was admitted gave the same pattern.

\textbf{The later source wins in both directions.} After the model answered, a false ``truncated'' header overturned
every correct admission of a clean pair (Nemotron 92/92, Haiku 64/64, Sonnet 99/99) and the true header nearly every
wrong refusal (Nemotron 11/11, Haiku 56/56, Sonnet 20/21). With the order reversed, the image overturned 98--100\% of
the wrong answers the false header had caused. In one prompt, by contrast, a false ``truncated'' header made all three
models refuse all 120 clean views: together, either objection sufficed.

\textbf{Not generic sycophancy.} ``Please confirm'' reversed 0 of 1,072 correct refusals of defects and 2 of 253
correct admissions. A bare claim without header values (``The pair is
complete'') reversed few Claude detections, so the header's numbers carry the effect; as a tool result rather than a
user message the header still reversed 52--99\%.

\textbf{Mitigation.} One sentence (``if the image and the header disagree, trust the image'') cut one-prompt override
to $\leq$2\% (Nemotron, Haiku, Sonnet; Gemma 5--8\%), at some cost in false refusal with a correct header (Nemotron
0\%$\to$12\%), and cut second-turn reversal to $\leq$9\%; it did not stop Haiku from following a false
``truncated'' header on clean pairs (95--96\%).

\textbf{Detection from the image varies.} Nemotron, Gemma, Haiku and Sonnet detected nearly all defects (SFR
$\leq$8\%); Llama-3.2-11B and MedGemma-1.5 did not discriminate (SFR $\geq$96\%); Muse-Glimmer-30B and Qwen3-VL-8B
refused nearly every admissible pair. 

\textbf{Real mismatches and pixels.} On 24 unedited NLST pairs with genuinely different coverage (median 1$-$z-IoU
0.13), of mismatches each model refused from the image, the false header overturned far more in the second turn than
in one prompt (aligned: Haiku 21/23 vs 5/23, Sonnet 10/21 vs 0/21, Nemotron 23/23 vs 19/23). A pixel-only check
missed 1 of 360 synthetic and 0 of 48 real defect views (AUC 0.98--0.99); robustness checks moved rates $\leq$3.3 points.

\section{Discussion and conclusion}
\label{sec:discussion}
A VLM that passes an image-only test can still give up defects it detected once contradicting metadata
arrives after its answer, while the same metadata given with the image is mostly handled conservatively. The effect needs the header's values, persists when the header arrives as pipeline (tool) output, and runs both ways.
Clinically, an agent that queries PACS after a visual check risks discarding correct detections, letting incomplete
series enter longitudinal volumetry unflagged. Plain recency does not fully explain this, as the same models behaved conservatively with both sources in one
prompt; the behaviour is consistent with evidence replacement, though we test behaviour, not mechanism.

\noindent\fbox{\parbox{\dimexpr\columnwidth-2\fboxsep-2\fboxrule}{\small
\textbf{Recommended validation for VLM input checks.} Before using a VLM to accept or reject imaging inputs, test it on
cases with known ground truth under: (1)~image only; (2)~image with contradicting metadata in the same prompt;
(3)~contradicting metadata after the model's own answer, as the pipeline would deliver it; and (4)~true metadata after a wrong answer, and a neutral
follow-up, as controls. Report missed
defects (SFR) and false refusals (FRR) separately in each. A check that passes (1) but fails (2) or (3) should not
see the metadata, or must be told it may be wrong; validate inputs outside the model where possible~\cite{medvigil}.}}

\textbf{Limitations.} One coronal plane per series, not the volume; native-view z-shifts are visible only through the
z labels, so that conflict is partly text vs text; headers are a synthetic one-line summary (raw fields tested on one
model); header-first was tested on the Claude models only; no GPT or Gemini model.

\section{Compliance with ethical standards}
This study was conducted retrospectively using only de-identified human subject data made available in open access
by the National Lung Screening Trial via the NCI Imaging Data Commons (CC BY 4.0). Ethical approval was not
required as confirmed by the license attached with the open access data.

\section{Acknowledgments}
Claude API costs were covered by the Liu Lab, School of Computing, University of Georgia. The author has no competing
interests.  \textit{Use of AI:} a generative AI assistant (Claude, Anthropic) wrote the evaluation and analysis code, executed
model API calls under the author's direction, and drafted and edited the text. The author chose the research direction,
directed which experiments were run and how results were framed, and reviewed all code, results and text, taking full
responsibility for the paper. All evaluated models, including Claude, used the same frozen prompts and analysis.

{\small
\bibliographystyle{IEEEbib}
\bibliography{refs}
}
\end{document}